\documentclass[11pt,a4paper]{article}
\usepackage[hyperref]{acl2018}
\usepackage{times}
\usepackage{latexsym}
\usepackage{times}
\usepackage{latexsym}
\usepackage{amsmath,amssymb,amsfonts}
\usepackage{algorithmic}
\usepackage{graphicx}
\usepackage{textcomp}
\usepackage{xcolor}
\usepackage{graphicx}
\usepackage{enumitem}
\usepackage{url}
\usepackage{booktabs}
\usepackage{soul}

\usepackage{url}

\aclfinalcopy % Uncomment this line for the final submission
\title{\textit{Helping each Other}: A Framework for Customer-to-Customer Suggestion Mining using a Semi-supervised Deep Neural Network }

\author{Hitesh Golchha, Deepak Gupta, Asif Ekbal, Pushpak Bhattacharyya \\
  Indian Institute of Technology Patna \\ Patna, India\\
  {\tt \{hitesh.me14, deepak.pcs16, asif, pb\}@iitp.ac.in} \\}

\date{}

\begin{document}
\maketitle
\begin{abstract}
 Suggestion mining is increasingly becoming an important task along with sentiment analysis. In today's cyberspace world, people not only express their sentiments and dispositions towards some entities or services, but they also spend considerable time sharing their experiences and advice to fellow customers and the product/service providers with two-fold agenda: helping fellow customers who are likely to share a similar experience, and motivating the producer to bring specific changes in their offerings which would be more appreciated by the customers. 
%The processing and natural language understanding of the recommendations, tips, feedback and suggestions emanating from users in online media thus has multiple use cases for both the fellow users as well as the product/service providers. 
In our current work, we propose a hybrid deep learning model to identify whether a review text contains any suggestion. The model %for classifying a review text into whether that 
employs semi-supervised learning to leverage the useful information from the large amount of unlabeled data. %algorithm to extract the mentions of suggestions from the customer reviews.
We evaluate the performance of our proposed model on a benchmark customer review dataset, comprising of the reviews of \textit{Hotel} and \textit{Electronics} domains. Our proposed approach shows the F-scores of $65.6\%$ and $65.5\%$ for the Hotel and Electronics review datasets, respectively. These performances are significantly better compared to %Our experimental results show the significant performance gains over the 
the existing state-of-the-art system. %the 
% We make a comparative analysis of the the performance of deep learning architectures enhanced with simple Semi Supervised Learning algorithm like self training against the existing state of the art,  \hl{We specifically only extract the mentions of suggestions from customers to other fellow customers.} 
%We make a comparative analysis of our system%\footnote{We specifically only extracted the mentions of suggestions from customers to other fellow customers} performance with the existing state of the art work.
%We achieved the F-scores of  $65.6\%$ and $65.5\%$ for the Hotel and Electronics review datasets respectively. Our experimental results show the significant performance gains over the existing state-of-the-art-work.
% , and we also perform a comprehensive analysis for demonstrating the superiority of our methods.
\end{abstract}

\section{Introduction}
The online platforms like social media websites, e-commerce sites of products and services, blogs, online forums and discussion forums etc. are very much attached today with our day-to-day lives.% of people.
The availability of the these information sharing platforms
%in today's online world 
has fueled the humans' desires to share one's opinions, emotions and sentiments with respect to the entities of all kinds: be it people, events, places, organizations, institutions, products, services, hobbies, games, movies, politics, technology etc. Generally people express their opinions in three ways: \textbf{(1)} through an independent piece of content writing \textbf{(2)} writing disposed towards a theme (such as a question in a community based question answering platform, or a topic in a discussion forum, or an entity in a product reviewing website/ e-commerce website) and  \textbf{(3)} conversational writings in the form of exchange of utterances in dialog systems/chats or comments for a post in social media/online forums.

Such opinions which exist in different forms and places, have often hidden in them the experiences of people, their subjective emotions and sentiments towards different aspects of different entities, as well as the intentions of advices and suggestions proposing some action in a prescribed way. Suggestion mining can be thought of as a subproblem of opinion mining, entrusted with the task of extracting mentions of suggestions from the unstructured texts. Suggestions in the domain of reviews can be generally of two kinds:
\begin{enumerate}[noitemsep]
\item \textbf{Customer to Companies:} These suggestions are directed from customers to the producers/service providers.
Customers provide companies with feedbacks, often expressing their contentment or complaining about their dissatisfaction with certain product features, services, processes or amenities. They provide detailed reasons and personal experiences for the same and offer alternative ideas for implementation. These kinds of suggestions are not only important as a tool for the companies to review their current offerings, but they are also a great source of ideas for new directions. 
\item \textbf{Customer to Customer:} These suggestions are provided from customers/users to the fellow customers/users.
Customers share their experiences in reviews, and provide tips and recommendations to the other customers. This is sometimes more than merely the % information is sometimes  which convey much more information than than 
information whether they like some specific attributes of the products or services. 
\end{enumerate}

\subsection{Motivation and Contributions}
There are several use cases of automated retrieval and natural language understanding for suggestion mining. Apart from their own experiences, understanding and knowledge, people depend on the online community to form their own opinions and readily look for suggestions and tips from the other customers. The extracted suggestions and tips are equivalent to a set of effective guidelines for the other customers before they make their own decisions. The fellow users can avail more information, and hence the decision taken would be better. This is often beyond the sense conveyed by aspect based sentiment analysis \cite{thet2010aspect,gupta2015pso,gupta-ekbal:2014:SemEval}. 
 
Suggestions and feedbacks are also an important component of the market survey performed by the companies to drive innovation, change and improvements. This task is a prerequisite to other nuanced tasks which include classifying the domain of the suggestion, identifying the other arguments of the suggestions (finding the entity towards whom the suggestion is directed, identifying the aspects regarding which %a tip is given or 
a suggestion has been made, finding the word boundaries of the suggestive expressions), and aggregation of such suggestions from multiple sources to comprehend a customer friendly summary.  

%\subsection{Contributions}
We summarize the contributions of our proposed work as follows:
\begin{itemize}[noitemsep]
\item We develop a linguistically motivated hybrid neural architecture to identify the review sentences that carry an intention of suggestion.% in reviews.
\item We employ semi-supervised learning (self-training) along with a deep learning based supervised classification approach. This gives us the opportunity to harness the treasure of  huge (unlabeled) data available in the form of customer reviews.
% for leveraging the useful information of unlabeled customer review data. 
To the best of our knowledge, this is the very first attempt in this direction to handle the target problem. %we are the first to have used semi supervised learning (self training ) to handle the given problem. 
\item Outperforming the current state-of-the-art customer-to- customer suggestion mining techniques and setting up a new state-of-the-art.
%\item Thorough analysis of the several linguistic features used to build the model.  
%\item A comprehensive study of the various errors encountered by the model and their intuitive explanations. 
\end{itemize}
% Our system outperforms better than the existing state of the art architectures in this task, which used LSTM and CNN as the prime components for classification.

% Moreover, in this work we propose the use of Self Learning along with supervised classification,  which is the first time semi supervised learning is being used for the task. This gives us the opportunity to harness the treasure of virtually unlimited data which is available for customer reviews.
\section{Related works}
%In this section we present a brief survey of the related works. 
%\subsection{Customer Suggestions Mining}
The field of suggestion classification and customer feedback analysis are relatively new in the area of Natural Language Processing (NLP) and Text Mining.
%new and the related literature is very fresh. 
Our work is most closely related to the prior research as reported in %done by Negi et. al.
\cite{negi2015towards,negi2016study}. %, and are an extension of it. 
In \cite{negi2015towards} authors defined the annotation guidelines for customer-to-customer suggestion mining. They trained a support vector machine (SVM) classifier over the features relevant for classification in the domains of hotels and electronics reviews. They used 
heuristic features, features extracted from the Part-of-Speech (PoS) tags, sequential pattern mining features, sentiment features and the features extracted from the dependency relations. 
\\
In their subsequent work, \cite{negi2016study} demonstrated the improved performance using Convolutional neural networks (CNN) \cite{kim2014convolutional} and Long short term memory (LSTM) \cite{hochreiter1997long} based deep learning architectures to solve this problem.
%for some of the publicly available datasets on suggestion mining (including the dataset of Negi et. al., 2015\cite{negi2015towards}) and some newly introduced ones. 
They experimented with both in-domain and cross-domain training data, and also compared their performance with a SVM based classifier trained with the same set of features similar to \cite{negi2015towards}. 

There are some other existing works for suggestion mining, beyond customer-to-customer suggestions. \cite{ngo2017identification} developed a binary classification model based on Maximum Entropy and CNN for filtering suggestion intents in Vietnamese conversational texts like posts, comments, reviews, messages chat and spoken texts. 
%They defined a different unit for identifying suggestion intents called functional units which are defined in ISO 24617-2 standard \footnote{ISO 24617-2:2012 provides a standard for dialogue annotations with concepts, a formal language (DiAML) and segmentation rules}, as the ``minimal stretch of communicative behavior that has one or more communicative functions." Such functional units may be helpful for the identification of suggestion intents in long and complex utterances.

Brun and Hagege \cite{brun2013suggestion} developed a feature-based suggestion mining system for the domain of product reviews. \cite{dong2013automated} performed suggestion mining on tweets of the customers regarding Microsoft Windows' phone. A model is proposed in \cite{wicaksono2013toward} which focused on extracting advices for the domains of travel using Hidden Markov Model (HMM) and Conditional Random Field (CRF).
The work as reported in \cite{gupta2017auto} focused on classifying the customer feedback sentences of users into six classes using deep learning based models.%- \textit{comment, request, bug, complaint, meaningless and undetermined} across four different
%languages. They developed both CNN and a CNN augmented RNN based classifiers for customer feedback analysis task.

Our proposed model differs from these existing works with respect to the problem addressed and the model developed. We have presented a very detailed comparison (in the experiments section) to the state-of-the-art system as reported in \cite{negi2015towards,negi2016study}. 

\section{Methodology}
In this section at first we discuss the various deep learning models and then semi-supervised model.
\subsection{Problem Definition}
Given a multi-sentence review $R$ having $N$ sentences $\{s_1, s_2, \ldots, s_N \}$ the task is to categorize each sentence $s_i$ into one of the classes $c\in C$ , where C =\{\textit{``suggestive"}, \textit{``non-suggestive} \}.
For a sentence $s$ with a sequence $w_{1}, w_{2}, \ldots , w_{n-1}, w_{n} $ of $n$ words, the associated suggestion class $c$ can be computed as:
\begin{equation}
c=argmax_{y} p(y|s) 
\end{equation}
where $y \in C$
% In this work, we extract mentions of explicitly conveyed suggestions of the latter kind, extending the work by \cite{negi2015towards} and using the same dataset. This retrieval problem has been designed as a sentence classification task by the authors.\\
% Given:  A sequence \textit{s} of \textit{n} word tokens comprising a sentence, \\
% Task: Assign this sequence a label of either 0 or 1, with label 1 indicating \textit{"suggestive"} sentence and a label 0 indicating \textit{"non-suggestive} sentence.

The following example review sentence, \textit{``Tip if you want a beach chair at the beach or pool, go there before 9 am or so and put your magazine or towel on your chair."} is a ``suggestion" intent directed towards a fellow customer. Here the expression of the intent is explicitly conveyed in the form of a review sentence with imperative mood\footnote{Imperative mood is a category or form of a verb which expresses a request or a command. For example, \textit{``Get ready"} }. The ``non-suggestive" sentences instead contain statements and facts (e.g. \textbf{(1)} \textit{``We stayed in the Westin Grand Berlin in July 2007."}) or expressions of one's sentiments (e.g. \textbf{(2)} \textit{``But the rooms are small and not very functional.")}. An interesting thing to note is that the second example has implicit suggestions for the fellow customers as well as the service provider (hotel owner). The other visiting customers are implicitly advised against renting the rooms of the hotel as they are small and have less utility. Moreover, this review sentence also consists of an implicit suggestion to the hotel owner to offer larger rooms to their customers, and also improve the functionalities that they provide. However in our work, we only deal with the suggestions which are very explicitly mentioned, and that too directed specifically to the fellow customers.
% . As our work is based on the annotations of \cite{negi2015towards} and thus the units of suggestion intents are sentences. 
\subsection{Proposed Deep Learning Model}
\begin{figure*}[h]
    \centering
    \includegraphics[width=5in]{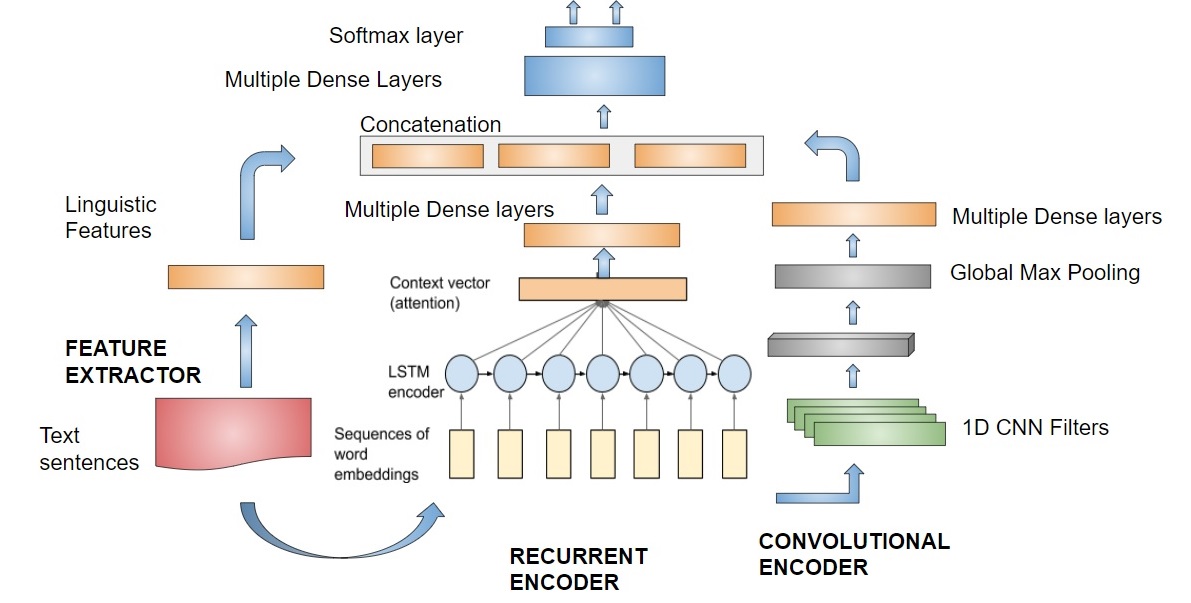}
    \caption{The proposed model architecture for customer suggestion mining}
    \label{fig:mesh1}
\end{figure*}

The customer-to-customer suggestion mining task requires recognizing specific syntactic and semantic constructions represented in texts. It should be able to capture the constructions representing imperative moods, and identify the patterns or phrases which are highly correlated with suggestive sentences in a review. It should also have a way for deep semantic understanding of text in order to disambiguate suggestions from the sentences which appear like suggestions on the surface.

We propose a hybrid model consisting of two deep learning based encoders designed to integrate different views or representations of the review sentences, and a linguistically motivated feature set. % in the form of linguistic knowledge. 
The information from the encoders along with linguistic knowledge are effectively combined with the help of a multi-layer perception (MLP) network. This is done to achieve higher abstraction necessary for a complex task like identifying the suggestive review sentence. Specifically, we use two different encoders, namely Convolutional Neural Network (CNN) and attention based Recurrent Neural Network (RNN). The effectiveness of CNN and RNN based encoder has been proven  in other NLP tasks \cite{C18-1042,tsd,GUPTA18.486,conll}. The CNN encoder uses multiple fully-connected over the convolution layer while the RNN encoder uses a LSTM layer with the attention \cite{raffel2015feed} followed by multiple fully-connected layers. An overview of the  architecture for suggestion mining is shown in Figure \ref{fig:mesh1}.

% Specifically the feature extractor encodes the relevant linguistic features, the convolutional encoder uses a few dense layers over one dimensional CNN filters while the recurrent encoder uses an LSTM layer with attention followed by some dense layers.

% The specific motivations for using these three encoders along with their descriptions are given below:
%We discuss the details in the following subsections: %The details of the linguistic features and both the encoders are discussed here on: 
\subsubsection{\textbf{Linguistic Features}} 
We use the following set of linguistic features in our model. We use slightly modified subset of features from \cite{negi2015towards} and similar to \cite{cicling} % component of our proposed architecture consists of the following linguistic features.  
% linguistic layer extracts a slightly modified  subset of features from \cite{negi2015towards} which includes:
%\begin{itemize}[noitemsep]
\paragraph{\textbf{Suggestive keywords}}: 
The suggestive keywords are usually associated with the texts containing actual suggestions. We use the following small set of suggestive keywords: \\
\textit{advice, suggest, may, suggestion, ask, warn, recommend, do, advise, request, warning, tip, recommendation, not, should, can, would, will}\\
%Some examples of the review sentences containing suggestive keywords are as follows:
%\begin{enumerate}
%\item I \textbf{would} not \textbf{recommend} this if you want a good nights sleep.
%\item It can be a point and shoot camera, but if you \textbf{do} not intend to  get into  its finer features, I'd \textbf{suggest} a less enthusiast featured choice.
%\item For prospective buyers, I would \textbf{warn} of the following 1 purchase a better pair of headphones as soon as possible, as ear buds sound terrible and are terribly uncomfortable this applies to ear buds.
%\end{enumerate}
A binary-valued feature is defined that checks whether the current word is one of the keywords or not
(1-presence, 0-absence). 

\paragraph{\textbf{N-gram features}}: 
%T%he n-gram feature is one of the very important features to categorize the review sentences.% into their respective category.
We extract the most frequent 300 unigrams, 100 bigrams, and  100 trigrams from the training set. These are then used as a bag of n-gram features.
\paragraph{ \textbf{Part-of-Speech (PoS) N-gram features}}:
%Similar to the n-gram features, 
We extract %the PoS n-gram features. We extract 
the most frequent PoS unigrams, bigrams and trigrams of size 50. These are then used as a bag of PoS n-grams features.

% We extracted most frequent top 50 Unigram, top 50 Bigrams, and top 50 Trigrams of the sentences, where the words have  been replaced by their Part of Speech Tags, and used them as features.
\paragraph{\textbf{Imperative mood features}}: Most of the suggestions containing sentences have imperative mood. We try to capture this phenomenon by introducing the features obtained from the dependency trees\footnote{We use spaCy dependency parser. For visualization, we used Stanford dependency parser}.
% which are essential in identifying such patterns. 
We use the following imperative mood features:
\begin{enumerate}
\item Base verb (VB) at the beginning of sentence or without nsubj arc: %Presence of the base form of word (PoS tag 'VB') as the beginning of any clause which does not contain any dependency arc with the label 'nsubj':
In many imperative sentences, the subject (denoted by nsubj) is absent, i.e. it implies to be the second person. Moreover, the clause containing the suggestive expression begins with the base form of the verb (denoted by VB). Hence, this does not have any dependency relation with nsubj. This feature is illustrated in Figure \ref{fig:mesh2}. 
%Presence of the base form of word (PoS tag 'VB') as the beginning of any clause which does not contain any dependency arc with the label 'nsubj'.
%In many imperative sentences, the subject is absent (and is implied to be the second person). Moreover, the clause containing the suggestive expression begins with the base form of the verb. The presence of this feature is illustrated in Figure \ref{fig:mesh2}. 
\begin{figure}[h]
\centering
\includegraphics[width=\linewidth]{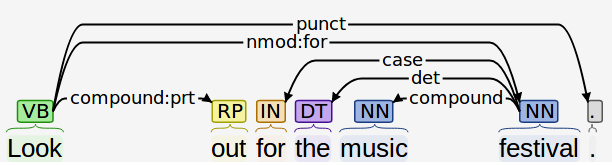}
\caption{Presence of `VB' without nsubj arc}
\centering
\label{fig:mesh2}
\end{figure}
\item `nsubj' dependency relation features: 
The pair of PoS tags of the words connected by the dependency arc `nsubj' is used as the bag of PoS feature. We describe the presence of this feature in Figure \ref{fig:dep1} and \ref{fig:dep2}.
\begin{figure}[h]
\centering
\includegraphics[width=\linewidth]{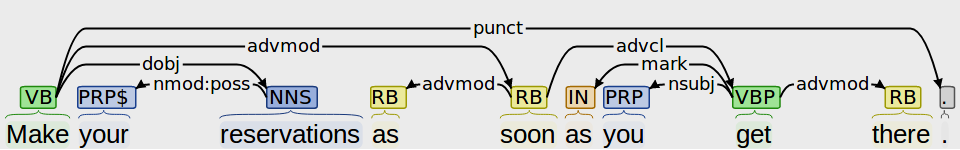}
\caption{nsubj dependency arc relations. From this dependency tree the extracted features are (VBP, PRP).
}
\label{fig:dep1}
\centering
\end{figure}

\begin{figure}[h]
\centering
\includegraphics[width=\linewidth]{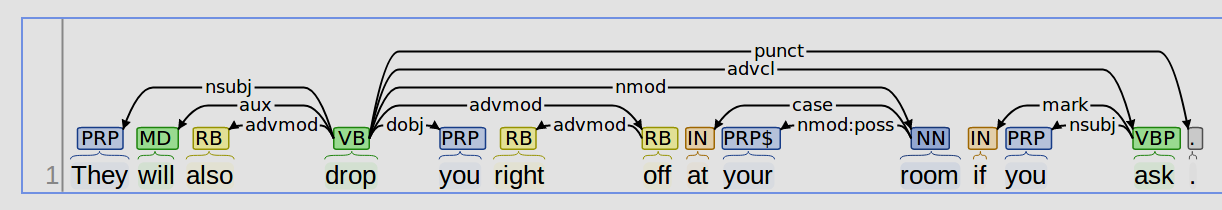}
\caption{nsubj dependency arc relations. Here, (VB, PRP) and (VBP, PRP) features are active.
}
\label{fig:dep2}
\centering
\end{figure}
\end{enumerate}
%\end{itemize}
% The linguistic features are the fed into two dense layers of 150 and 25 neurons, each having ReLU activations and Dropouts \cite{srivastava2014dropout} of 0.2. \\
This set of linguistic features are fed into a multilayer perceptron having two hidden layers of size $150$ and $25$, respectively.
\subsubsection{\textbf{Recurrent and CNN Encoders}}
The words in the sequence $\{w_1, w_2 \ldots w_n \}$ from a given review sentence $s$ are mapped to their corresponding word vectors $\{x_1, x_2 \ldots x_n\}$. The word embeddings are obtained through the publicly available\footnote{https://nlp.stanford.edu/projects/glove/} GloVe word embeddings\cite{pennington2014glove} of dimension $300$ and trained on the Common Crawl. 
% Prior to the Recurrent and Convolutional encoders, the words in input sentence are represented by 300 dimensional word embeddings pretrained using the Glove algorithm \cite{pennington2014glove}, and a dropout of 0.25 is applied on them. 

The recurrent encoder uses a LSTM network (hidden size 64) over the embedded sequences and it then applies an internal attention over the hidden states.

The LSTM network is able to process the sentence as a sequence, with the ability to capture long term dependencies. Thus the hidden layers can efficiently perform composition over the local context, and help to identify patterns which are found in suggestive sentences. The attention mechanism then finds salient contexts and aggregates the important ones to build the context vector. The motivation for using attention stems from the fact that suggestive expressions can be identified in a short span of text within the sentence and the attention can effectively attend to those specific contexts encoded by LSTM. The Attention layer is followed by dense layers with 150, and 25 neurons, each having ReLU activations and a dropout value of 0.2.  
%\subsubsection{\textbf{Convolutional Encoder}}

The convolutional layer applies 250 one dimensional CNN filters of size 5 over the embeddings. The global max pooling is applied separately for the feature map obtained from each filter, and it helps to identify the presence of the n-gram feature corresponding to that feature in the sentence. The following dense layer with 250 neurons (ReLU activation and 0.75 dropout) helps to non-linearly compose multiple such features, thus giving itself an opportunity to learn a more diverse set of features.  

\subsubsection{\textbf{Hybrid Model}}
The extracted linguistic features, the recurrent encoder representation and the convolutional encoder representation are concatenated (into a feature set $p$) and fed to a fully-connected layer with two neurons, followed by softmax activation. The softmax layer outputs the probability of the given review sentence being \textit{suggestive} or \textit{non-suggestive}.  %This gives the neural network an opportunity to design higher level features composed of these lower level features. 
%Ideally, the softmax layer normalizes the output to a probability distribution over the two classes- \textit{suggestions} and \textit{non-suggestions}. 
The probability that the output class $\hat{y}$ is $i$ given the sentence $s$ and parameters $\theta$ is computed as: 
\begin{equation}
\begin{split}
P(\hat{y}=i|s, \theta)& = softmax_i(p\textsuperscript{T}w_i+z_i)\\
&=\frac{\mathrm{e}^{p^{T}w_i+z_i}}{ \sum_{k=1}^{K}\mathrm{e}^{p^{T}w_k+z_k}}
\end{split}
\end{equation}
where $z_k$ and $w_k$ are the bias and weight vector of the $k$\textsuperscript{th} labels, $p$ is the concatenated feature set, and $K$ is the number of total classes (i.e. 2). $\theta$ is the set of all the parameters of the model. The system predicts the most probable class.
\subsection{Semi-supervised Model}
%The network parameters are trained in both supervised and semi-supervised manners to make a comparative analysis. 
%The reimplementation of models of Sapna et. al.,2016 (\cite{negi2016study})  are trained in the same manner, for uniformity in comparisons.

Semi-supervised learning makes use of %is the domain of machine learning concerned which uses 
both labeled (small) and unlabeled (huge) data for designing a more efficient classifier, as compared to the traditional supervised learning. We utilize self-training algorithm \cite{zhu2006semi}, also known as bootstrapping, which can be flexibly used as a wrapper over any supervised learning algorithm. We use our hybrid model for this semi-supervised learning. %\hl{Thus, we use it with our hybrid model.}
%({\bf comment: mention here which algorithm we used? Is it LSTM?})

\begin{figure}[h]
\centering
\includegraphics[width=2.5in]{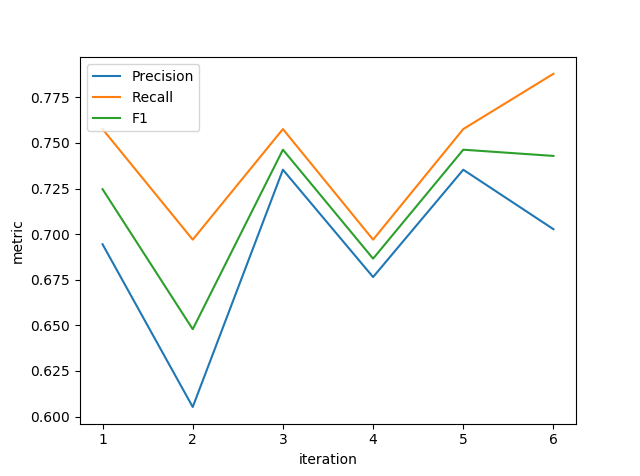}
\caption{ Scores on the validation set during self-training: Hotel domain.}
\centering
\label{fig:selftrain_hotel}
\end{figure}

\begin{figure}[h]
\centering
\includegraphics[width=2.5in]{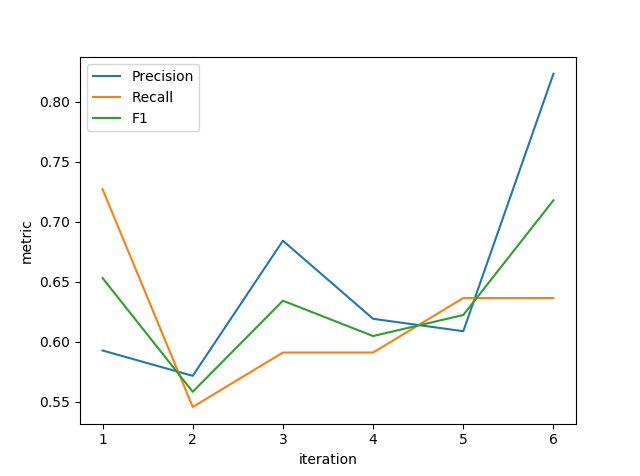}
\caption{ Scores on the validation set during self-training: Electronics domain.}
\centering
\label{fig:selftrain_electronics}
\end{figure}

In self-training, we iteratively train a classifier enhancing each time the original training dataset with newly labeled instances. At the end of each iteration, the classifier is made to predict on the unlabeled dataset and $100$ most confidently predicted instances of each class is added to the training data, with the predicted labels as the true labels. For self-training, a methodology similar to early stopping is applied, with a maximum of six iterations. We stop the iteration %Further iterations are stopped 
when the F1-Score on the validation data \footnote{A part of the training set was used for validation.} does not improve over the existing best model in consecutive three iterations, saving only the best performing model for testing. For example in Fig. \ref{fig:selftrain_hotel}, the training terminates after the 6th iteration, and the model trained in the 3rd iteration is chosen for the final evaluation. Effect of adding unlabeled data to training for the electronics domain is depicted in Figure \ref{fig:selftrain_electronics}.

%In \textbf{supervised} setting, 
For this semi-supervised setting, the cross-entropy error is minimized using the Adam Optimizer, and the training is stopped as the validation loss stops decreasing (early stopping). Because of the class imbalance (cf. Table \ref{tab:table1}), the loss function weighs the loss for the positive class instances 10 times more than the loss for the  negative class instances. %The other configurations like the loss function, classifier, optimizer, early stopping methodology are the same as that in supervised settings. 
%We keep the hidden vector size of the LSTM to be $64$ in our experiment. The optimal filter size is found to be $\{5\}$ and the number of filters is set to be $250$. We use ReLu as the activation function in our experiment. 
All the other configurations are similar to the supervised setting\footnote{Models are optimized based on the validation set, a part of the training set}.
%. , and We obtain these values via the 5-fold experiments.
\section{Dataset and Experiments}
In this section we present the datasets, experimental results and the necessary analysis. 
\subsection{Dataset}
We conduct experiments on the dataset created by \cite{negi2015towards}). The dataset comprises of the sentences of reviews taken from two domains, \textit{viz.} Hotels and Electronics. The dataset was annotated as `suggestive' and `non-suggestive'. 

The hotel reviews in \cite{negi2015towards} are a subset of the TripAdvisor reviews annotated by   \cite{wachsmuth2014review}), with the sentiment polarity classes of positive, negative, neutral and conflicting. The electronics dataset was originally annotated by  \cite{hu2004mining}) with the sentiment labels, and \cite{negi2015towards} extended it for suggestion mining. The dataset consists of 7534 sentences from the hotel reviews and 3782 sentences from the reviews of electronic items. For semi-supervised learning experiments, we obtain the complete dataset from  \cite{wachsmuth2014review} for hotels. We segment these reviews into 21328 sentences in total. For the electronics domain, we use the Amazon reviews obtained from the electronics segment of \cite{He2016UpsAD} as the unlabeled data. The first 50,000 sentences extracted from the reviews were chosen for the experiments.

\begin{table}[h!]
 \centering  
    \resizebox{2.5in}{!}{%

    \begin{tabular}{l|c|c}
      \toprule % <-- Toprule here
       & Hotel  & Electronics \\
       &  reviews &  reviews\\
      
      \midrule % <-- Midrule here
      Class = 1 \textit{Suggestive} & 407 & 273\\
      Class = 0 \textit{Non - Suggestive} & 7127 & 3509\\
      Total     &7534 & 3782\\
      \midrule
      Class1 : Class0 &1:17.5 & 1:12.9\\
      \bottomrule % <-- Bottomrule here
    \end{tabular}%
}
    \caption{Dataset statistics (on the sentence level)}
    \label{tab:table1}

\end{table}
Instances of suggestions and tips form a relatively small percentage of the total review sentences, and this is reflected in the class distribution of the labeled dataset. The number of instances is not enough for very deep  architectures. Statistics of the datasets are presented in Table \ref{tab:table1}.
\subsection{Results and Analysis}
We re-implement the LSTM and CNN architectures proposed in \cite{negi2016study} to construct our baselines. We re-implement this state-of-the-art system with the common training methodologies as ours. %The LSTM with attention is developed using recurrent encoder followed by a fully connected layer with two neurons, and then a softmax layer.
Detailed evaluation results are demonstrated in Table \ref{tab:table2}. %We also show the evaluation results of the feature based SVM model in \cite{negi2015towards}. The classifier was trained with heuristic features, generic features extracted from n-grams of PoS tags, and the special features like sequential pattern mining (which seek to extract imperative mood specific expressions), sentiment features and subject specific features (extracted from the dependency relations).
% as one of our baselines. 

\begin{table*}[h!]
  \begin{center}
\resizebox{4in}{!}{%
    
    \begin{tabular}{l|c|c|c|c|c|c}
      \toprule % <-- Toprule here
       Model &  &Hotel & &   &Electronics & \\
         \midrule % <-- Midrule here
      & Precision & Recall & F1 &  Precision & Recall & F1\\
      \midrule % <-- Midrule here
      
      CNN            & 0.560   & 0.641  & 0.598 &    0.586 & 0.615 & 0.600  \\
      LSTM        & 0.511   & 0.624  & 0.562 &   0.582 & 0.644 & 0.611\\
      LSTM + Attention & 0.494   & \textbf{0.769}  & 0.602    & 0.543 & \textbf{0.699} & 0.611\\
      \citet{negi2015towards} & 0.580   & 0.512  & 0.567    & \textbf{0.645} & 0.621 & 0.640\\
      Proposed Hybrid  & 0.593   & 0.703  & 0.643 &    0.587 & 0.660 & 0.621\\
      Proposed Hybrid +   \\
       Self Training & \textbf{0.639}   & 0.673  & \textbf{0.656}    & 0.634 & 0.677 & \textbf{0.655} \\

      \bottomrule % <-- Bottomrule here
    \end{tabular}%
}
        \caption{Macro average evaluation results on 5-fold cross validation. Results of CNN and LSTM are based on the reimplementation of \cite{negi2016study}}
        \label{tab:table2}
  \end{center}
\end{table*}

\begin{table*}[h!]
  \begin{center}
	\resizebox{4in}{!}{%
   
    \begin{tabular}{l|c|c|c|c|c|c}
      \toprule % <-- Toprule here
       Model &  &Hotel & &   &Electronics & \\
         \midrule % <-- Midrule here
      & Precision & Recall & F1 &  Precision & Recall & F1\\
      \midrule % <-- Midrule here
      
      Hybrid            & 0.593   & 0.703  & 0.643 &    0.587 & 0.660 & 0.621  \\
      Hybrid - CNN encoder        & 0.585   & 0.696  & 0.636 &   0.542 & 0.721 & 0.618\\
      Hybrid - RNN encoder  & 0.636   & 0.641  & 0.638    & 0.586 & 0.644 & 0.614\\
      Hybrid - Linguistic encoder  & 0.554   & 0.626  & 0.588 &  0.615   & 0.633 & 0.624 \\

\midrule
      \bottomrule % <-- Bottomrule here
    \end{tabular}%
}
        \caption{Macro average evaluation on 5-fold cross validation for the ablation study of different component models }
         \label{tab:table3}
  \end{center}
\end{table*}

%The primary aim of the systems was to extract the suggestions present in the text, which are sparsely distributed amongst the review sentences, thus making precision, recall, and F1 score of the suggestion class an important metric for measuring the quality of the developed systems. 

%The LSTM and the CNN based models are considered as the baselines in our task. 
The LSTM is capable of handling long term dependencies and that may be attributed to its better performance against CNN for the domain of electronics where the average sentence length is relatively longer. The model based on LSTM achieves the F1 scores of 0.562 and 0.611 for the hotel and electronics datasets, respectively. %The CNN filters with max pooling are able to effectively search for n-gram features invariant to translation, and thus the 
CNN based model also demonstrates comparative performance with F1 scores of 0.598 and 0.600 for the two domains, respectively. Introducing attention to the LSTM model was found to be effective with reasonable performance improvement. %The addition of attention was found to be effective with performance improvement of the LSTM system, and because of its presence, 
Because of attention the system could attend to specific regions of the input sentence which had patterns similar to that of suggestive sentences, encoded by its query vector. The system with attention shows the best recall of 76.9\% for the hotel reviews and 69.9\% for the electronics reviews, establishing our claim about its ability. It achieves the F1 scores of 0.602 and 0.611 for the two domains, respectively. 

Among the different architectures, the proposed hybrid model is found to be the best performing one with F1 scores of 0.643 and 0.621 for the two domains, respectively. Each of the encoders provides a different representation and features of the input, and the dense layers are able to combine them in an effective way. % to predict the correct answer. 
We also remove different encoders-one after other-from the proposed hybrid system to analyze the importance of each. Ablation studies of these models are reported in Table \ref{tab:table3}. For hotel reviews the order of importance of feature encoders are: \textit{Linguistic encoder \textgreater  CNN Encoder \textgreater  RNN Encoder}. For electronics reviews the importance of model encoders are: \textit{RNN Encoder  \textgreater  CNN Encoder \textgreater  Linguistic Encoder}. Effectively, different representations of the review sentences and the corresponding features are indeed important for the classification task. 

The use of self-training further improves the precision of the proposed hybrid model because it conservatively adds high confidence predictions obtained from the unlabeled data to the training data in each iteration. Inclusion of the 	`suggestion' class examples into training helped in reducing the class imbalance, which leads to the improved recall scores for the positive class. Augmentation of new data also added more lexical variability for the system to learn. This, in turn, helps for better classification with %to distinguish better between suggestive and non-suggestive instances. 
%All these attributed to 
the improved F1 scores of 0.656 and 0.655 for the hotel and electronics domains, respectively. 
%({\bf comment: pl mention how many new samples we have added after self training for both the domains})
The self training runs for a mean of 3.2 (SD = 0.75) iterations for the hotel domain, and 3.6 iterations (SD = 1.62) for the electronics domain. Thus, the expected number of unlabeled sentences added are 640 and 720, respectively. %, for these two domains.  
Our proposed system clearly performs better than the state-of-the-art model \cite{negi2015towards} with the increments of 8.9 and 1.5 F1 score points for the hotel and electronics domain, respectively. Please note that the SVM based model was trained with a diverse and rich feature set. Statistical T-test show the performance improvement as significant. 
%({\bf comment: pl report the statistical significance test here to show that our model is really better. you can try t-test})

\section{Error Analysis}
\begin{figure}[h]
    \centering
    \includegraphics[width=2
    in]{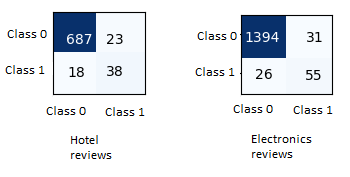}
    \caption{Confusion matrix on test set using Hybrid+Self- training. Here, 1: suggestive and 0: non-suggestive}
    \label{fig:cf1}
     \centering
\end{figure}

In order to understand the behaviors of our proposed model, we perform error analysis-both quantitatively and qualitatively. For quantitative analysis we depict the confusion matrix in Figure \ref{fig:cf1}.
%We perform detailed analysis of the outputs produced by our proposed model.
Our closer analysis reveals that a lot of electronics reviews are slightly longer and more complex than the hotel reviews, making it slightly harder to predict despite having slightly more balanced class distribution. Moreover, the presence of only 273 reviews of class 1 in all (about 218 reviews in training), is too small for the architectures to effectively model. %Overall, we observe that the systems have learned the ability to identify the sentences which contain suggestive terms and also sentences which are imperative in nature. 

We provide more detailed analysis with the actual examples. At first we describe the phenomena where the instances are incorrectly predicted as suggestions (i.e. false positive cases):
\begin{itemize}[noitemsep]
\item Many of the false positives are of imperative mood, but do not contain any suggestion towards any entity or product. For e.g.:\\
\textit{ Forget the fact that it will probably take me a year to figure out all the features this camera has to offer.}
\item Sometimes a user shares his/her own experience(s) but in the form of a second person, thus confusing the machine to predict as a suggestive sentence. For e.g.\\
\textit{“You book a top floor, you get first floor, you booked a suite, and got a room...you go out to your balcony to relax...and someone from a top floor....,which you reserved, has just spit on the back of your head.”}
 
\item Many of the sentences consist of objects in the second person, but the sentences are not imperative in mood. Such errors are more common in the CNN based model, but are lesser in the proposed hybrid system that makes use of self-training.
For e.g.\\ 
\textit{“If we find some great cheap places we will share it with you.”} \\
\textit{``Sentence: very comfortable camera , easy to use , and the best digital photos you re going to get at this price"}

\item LSTM model sometimes incorrectly predicts those review texts where tokens with VB PoS tags appear. %The review texts where words with %If a word with the tag 
%VB PoS tag appear %is present in between the sentence, such examples 
%are predicted as suggestive by the LSTM based model. 
This happens because the sentence appears to be similar to a suggestive sentence that also starts from that particular word having VB PoS tag.
For e.g.\\ \textit{You need the storage to hold a decent amount of shots at 4 megapixel resolution} \\ might be confused with\\  \textit{Hold a decent amount of ....}

\item Some of the false positives are actually suggestions which appear to be wrongly labeled in the original dataset. %mis-annotated.
\textit{“ We would definitely recommend this hotel to our friends.”}
 
 \item Suggestions against a product/service which are sarcastic in nature have been annotated as non-suggestive but are difficult for our system to differentiate from the usual suggestions.
 
\textit{“I recommend this hotel only if you don't mind blithely throwing money around, and if you bring your own towels”}

\end{itemize}
We also show here few examples that contribute to the false negatives:
\begin{itemize}
\item When the sentences are very long, and only a clause of the text belongs to the imperative mood, it is missed by  even the best system. 
For e.g. \textit{``The battery lasts very long when playing music, but writing files to the player drains the battery fast , so you need to have it plugged into an outlet when sending files. "} 
%\par One solution may be classifying clauses separately-which we will explored in our future work.
\item  Sometimes two sentences are clubbed together into one when the end marker is missing. In such scenarios, one of the sentences is suggestive and the other is not. In these cases the system predicts the sentence as non-suggestive.
For e.g. \\
\textit{``My only suggestion is to get a lens protector to help protect the shooting lens  the lens coating will wear out after so many clean wipes  and I m getting the those  52 mm adapter and uv lens filter  at lensmateonline.com ."}

It becomes more tricky for the machine if the remaining part of the sentence contains multiple occurrences of first person pronoun. For e.g. \textit{``You have to press the buttons hard and frequently I end up pressing enter when I meant to scroll ."}
\end{itemize}
From qualitative analysis we observe that systems have learned the ability to identify the sentences with suggestive terms and also the sentences which are imperative in nature. %We categorize the typical errors into few groups as described below: %and describe below: % Below we describe the typical error cases with appropriate groupings. 
We believe that many of these errors can be reduced to a greater extent by increasing the size of the training data. % which is small with respect to the enormous expressive power of the current deep learning systems. 
With sufficient data, systems would be able to learn to better model the input, extract more relevant features, and be able to reason better about the differences between the suggestive sentences and sentences which look like suggestions.
\section{Conclusion and Future Work}
In this paper, we have proposed a hybrid deep learning model for the task of suggestion mining by incorporating richer and diverse representations of the inputs. 
We have also used self-training algorithm, which even improved the performance of the hybrid model, opening up more opportunities for the use of semi-supervised learning for this task. Experiments on  benchmark datasets show that we obtain superior performance over the existing state-of-the-art system. % architectures that have been used in this domain.
In the future, we would like extend our work to other semi-supervised learning algorithms.
%, % which can act as wrapper over supervised deep learning architectures, 
%specifically in line of the work of \cite{bui2017neural} where they regularize neural networks using an objective function which is motivated by the label propagation algorithm in semi-supervised learning. %This has been shown to work effectively in several domains, including document and semantic intent classification.
\bibliography{acl2018}

\begin{thebibliography}{23}
\expandafter\ifx\csname natexlab\endcsname\relax\def\natexlab#1{#1}\fi

\bibitem[{Brun and Hagege(2013)}]{brun2013suggestion}
Caroline Brun and Caroline Hagege. 2013.
\newblock Suggestion mining: Detecting suggestions for improvement in users'
  comments.
\newblock \emph{Research in Computing Science}, 70:199--209.

\bibitem[{Dong et~al.(2013)Dong, Wei, Duan, Liu, Zhou, and
  Xu}]{dong2013automated}
Li~Dong, Furu Wei, Yajuan Duan, Xiaohua Liu, Ming Zhou, and Ke~Xu. 2013.
\newblock The automated acquisition of suggestions from tweets.
\newblock In \emph{AAAI}.

\bibitem[{Gupta et~al.(2018{\natexlab{a}})Gupta, Ekbal, and
  Bhattacharyya}]{GUPTA18.486}
Deepak Gupta, Asif Ekbal, and Pushpak Bhattacharyya. 2018{\natexlab{a}}.
\newblock {A Deep Neural Network based Approach for Entity Extraction in
  Code-Mixed Indian Social Media Text}.
\newblock In \emph{Proceedings of the Eleventh International Conference on
  Language Resources and Evaluation ({LREC 2018})}, Miyazaki, Japan. European
  Language Resources Association (ELRA).

\bibitem[{Gupta et~al.(2018{\natexlab{b}})Gupta, Kohail, and
  Bhattacharyya}]{cicling}
Deepak Gupta, Sarah Kohail, and Pushpak Bhattacharyya. 2018{\natexlab{b}}.
\newblock Combining graph-based dependency features with convolutional neural
  network for answer triggering.
\newblock \emph{arXiv preprint arXiv:1808.01650}.

\bibitem[{Gupta et~al.(2017)Gupta, Lenka, Bedi, Ekbal, and
  Bhattacharyya}]{gupta2017auto}
Deepak Gupta, Pabitra Lenka, Harsimran Bedi, Asif Ekbal, and Pushpak
  Bhattacharyya. 2017.
\newblock \href {http://aclweb.org/anthology/I17-4031} {Iitp at ijcnlp-2017
  task 4: Auto analysis of customer feedback using cnn and gru network}.
\newblock In \emph{Proceedings of the IJCNLP 2017, Shared Tasks}, pages
  184--193. Asian Federation of Natural Language Processing.

\bibitem[{Gupta et~al.(2018{\natexlab{c}})Gupta, Lenka, Ekbal, and
  Bhattacharyya}]{conll}
Deepak Gupta, pabitra Lenka, Asif Ekbal, and Pushpak Bhattacharyya.
  2018{\natexlab{c}}.
\newblock Uncovering code-mixed challenges: A framework for linguistically
  driven question generation and neural based question answering.
\newblock In \emph{Proceedings of 22nd International Conference on
  Computational Natural Language Learning . ({CoNLL 2018})}. Association for
  Computational Linguistics (Accepted).

\bibitem[{Gupta et~al.(2018{\natexlab{d}})Gupta, Pujari, Ekbal, Bhattacharyya,
  Maitra, Jain, and Sengupta}]{C18-1042}
Deepak Gupta, Rajkumar Pujari, Asif Ekbal, Pushpak Bhattacharyya, Anutosh
  Maitra, Tom Jain, and Shubhashis Sengupta. 2018{\natexlab{d}}.
\newblock \href {http://aclweb.org/anthology/C18-1042} {Can taxonomy help?
  improving semantic question matching using question taxonomy}.
\newblock In \emph{Proceedings of the 27th International Conference on
  Computational Linguistics ({COLING 2018})}, pages 499--513. Association for
  Computational Linguistics.

\bibitem[{Gupta and Ekbal(2014)}]{gupta-ekbal:2014:SemEval}
Deepak~Kumar Gupta and Asif Ekbal. 2014.
\newblock \href {http://www.aclweb.org/anthology/S14-2053} {{IITP}:
  {S}upervised {M}achine {L}earning for {A}spect based {S}entiment {A}nalysis}.
\newblock In \emph{{P}roceedings of the 8th {I}nternational {W}orkshop on
  {S}emantic {E}valuation ({S}em{E}val 2014)}, pages 319--323, Dublin, Ireland.
  Association for Computational Linguistics and Dublin City University.

\bibitem[{Gupta et~al.(2015)Gupta, Reddy, Ekbal et~al.}]{gupta2015pso}
Deepak~Kumar Gupta, Kandula~Srikanth Reddy, Asif Ekbal, et~al. 2015.
\newblock \href
  {https://link.springer.com/chapter/10.1007/978-3-319-19581-0_20}
  {P{S}{O}-{A}{S}ent: {F}eature {S}election using {P}article {S}warm
  {O}ptimization for {A}spect {B}ased {S}entiment analysis}.
\newblock In \emph{{I}nternational {C}onference on {A}pplications of {N}atural
  {L}anguage to {I}nformation {S}ystems {(NLDB-2015)}}, pages 220--233.
  {S}pringer.

\bibitem[{He and McAuley(2016)}]{He2016UpsAD}
Ruining He and Julian McAuley. 2016.
\newblock Ups and downs: Modeling the visual evolution of fashion trends with
  one-class collaborative filtering.
\newblock In \emph{WWW}.

\bibitem[{Hochreiter and Schmidhuber(1997)}]{hochreiter1997long}
Sepp Hochreiter and J{\"u}rgen Schmidhuber. 1997.
\newblock Long short-term memory.
\newblock \emph{Neural computation}, 9(8):1735--1780.

\bibitem[{Hu and Liu(2004)}]{hu2004mining}
Minqing Hu and Bing Liu. 2004.
\newblock Mining and summarizing customer reviews.
\newblock In \emph{Proceedings of the tenth ACM SIGKDD international conference
  on Knowledge discovery and data mining}, pages 168--177. ACM.

\bibitem[{Kim(2014)}]{kim2014convolutional}
Yoon Kim. 2014.
\newblock \href {http://www.aclweb.org/anthology/D14-1181} {Convolutional
  neural networks for sentence classification}.
\newblock In \emph{Proceedings of the 2014 Conference on Empirical Methods in
  Natural Language Processing (EMNLP)}, pages 1746--1751, Doha, Qatar.
  Association for Computational Linguistics.

\bibitem[{Maitra et~al.(2018)Maitra, Sengupta, Gupta, Pujari, Ekbal,
  Bhattacharyya, Maitra, Abhisek, and Jain}]{tsd}
Anutosh Maitra, Shubhashis Sengupta, Deepak Gupta, Rajkumar Pujari, Asif Ekbal,
  Pushpak Bhattacharyya, Anutosh Maitra, Mukhopadhyay Abhisek, and Tom Jain.
  2018.
\newblock Semantic question matching in data constrained environment.
\newblock In \emph{Proceedings of the 21st International Conference on Text,
  Speech and Dialogue ({TSD-2018})}, pages 499--513.

\bibitem[{Negi et~al.(2016)Negi, Asooja, Mehrotra, and
  Buitelaar}]{negi2016study}
Sapna Negi, Kartik Asooja, Shubham Mehrotra, and Paul Buitelaar. 2016.
\newblock A study of suggestions in opinionated texts and their automatic
  detection.
\newblock In \emph{Proceedings of the Fifth Joint Conference on Lexical and
  Computational Semantics}, pages 170--178.

\bibitem[{Negi and Buitelaar(2015)}]{negi2015towards}
Sapna Negi and Paul Buitelaar. 2015.
\newblock Towards the extraction of customer-to-customer suggestions from
  reviews.
\newblock In \emph{Proceedings of the 2015 Conference on Empirical Methods in
  Natural Language Processing}, pages 2159--2167.

\bibitem[{Ngo et~al.(2017)Ngo, Pham, Takeda, Pham, and
  Phan}]{ngo2017identification}
Thi-Lan Ngo, Khac~Linh Pham, Hideaki Takeda, Son~Bao Pham, and Xuan~Hieu Phan.
  2017.
\newblock On the identification of suggestion intents from vietnamese
  conversational texts.
\newblock In \emph{Proceedings of the Eighth International Symposium on
  Information and Communication Technology}, pages 417--424. ACM.

\bibitem[{Pennington et~al.(2014)Pennington, Socher, and
  Manning}]{pennington2014glove}
Jeffrey Pennington, Richard Socher, and Christopher Manning. 2014.
\newblock Glove: Global vectors for word representation.
\newblock In \emph{Proceedings of the 2014 conference on empirical methods in
  natural language processing (EMNLP)}, pages 1532--1543.

\bibitem[{Raffel and Ellis(2015)}]{raffel2015feed}
Colin Raffel and Daniel~PW Ellis. 2015.
\newblock Feed-forward networks with attention can solve some long-term memory
  problems.
\newblock \emph{arXiv preprint arXiv:1512.08756}.

\bibitem[{Thet et~al.(2010)Thet, Na, and Khoo}]{thet2010aspect}
Tun~Thura Thet, Jin-Cheon Na, and Christopher~SG Khoo. 2010.
\newblock Aspect-based sentiment analysis of movie reviews on discussion
  boards.
\newblock \emph{Journal of information science}, 36(6):823--848.

\bibitem[{Wachsmuth et~al.(2014)Wachsmuth, Trenkmann, Stein, Engels, and
  Palakarska}]{wachsmuth2014review}
Henning Wachsmuth, Martin Trenkmann, Benno Stein, Gregor Engels, and Tsvetomira
  Palakarska. 2014.
\newblock A review corpus for argumentation analysis.
\newblock In \emph{International Conference on Intelligent Text Processing and
  Computational Linguistics}, pages 115--127. Springer.

\bibitem[{Wicaksono and Myaeng(2013)}]{wicaksono2013toward}
Alfan~Farizki Wicaksono and Sung-Hyon Myaeng. 2013.
\newblock Toward advice mining: Conditional random fields for extracting
  advice-revealing text units.
\newblock In \emph{Proceedings of the 22nd ACM international conference on
  Information \& Knowledge Management}, pages 2039--2048. ACM.

\bibitem[{Zhu(2006)}]{zhu2006semi}
X~Zhu. 2006.
\newblock Semi-supervised learning literature survey, department of computer
  sciences, university of wisconsin at madison, madison.
\newblock Technical report, WI, Technical Report 1530. http://pages. cs. wisc.
  edu/\~{} jerryzhu/pub/ssl\_survey. pdf.

\end{thebibliography}
\bibliographystyle{acl_natbib}

\end{document}